\documentclass[lettersize,journal]{IEEEtran}
\usepackage{amsmath,amsfonts}
\usepackage{algorithmic}
\usepackage{algorithm}
\usepackage{array}
\usepackage[caption=false,font=normalsize,labelfont=sf,textfont=sf]{subfig}
\usepackage{textcomp}
\usepackage{stfloats}
\usepackage{url}
\usepackage{verbatim}
\usepackage{graphicx}
\usepackage{cite}
\usepackage{booktabs,multirow}
\usepackage{tabularray}
\usepackage{color}
\usepackage{multirow}
\usepackage{colortbl}
\usepackage{graphicx}
\usepackage[table,xcdraw]{xcolor}
\usepackage{hyperref}
\usepackage{cleveref}
\UseRawInputEncoding
\begin{document}

\title{CCTNet: A Circular Convolutional Transformer Network for LiDAR-based Place Recognition Handling Movable Objects Occlusion}

\author{Gang~Wang,~\IEEEmembership{}Chaoran~Zhu,~\IEEEmembership{}Qian~Xu,~\IEEEmembership{}Tongzhou~Zhang,~\IEEEmembership{}Hai~Zhang,~\IEEEmembership{Member,~IEEE, }\\Xiaopeng~Fan,~\IEEEmembership{Senior~Member,~IEEE, }Jue~Hu~\IEEEmembership{}\vspace{-10mm}
\thanks{
This work is supported by the Jilin Scientific and Technological Development Program (20210401145YY), the Changsha Automobile Innovation Research Institute (CAIRIZT20220101), the National Key Research and Development Program of China (2023YFE0197800), and the science foundation of national key laboratory of science and technology on advanced composites in special environments (JCKYS2023603C014). (Corresponding authors: Tongzhou Zhang)}
\thanks{Gang Wang, Chaoran Zhu and Tongzhou Zhang are with the College of Computer Science and Technology, Jilin University, Changchun 130012, P.R.China (e-mail: gangwang@jlu.edu.cn; zhucr22@mails.jlu.edu.cn; tzzhang20@mails.jlu.edu.cn).}
\thanks{Qian Xu, China North Vehicle Research Institute (NOVERI), Beijing, 100072, P. R. China (e-mail: quincyxu@hotmail.com).}
\thanks{Hai Zhang is with the Centre for Composite Materials and Structures, Harbin Institute of Technology, Harbin 150001, P.R.China (e-mail: hai.zhang@hit.edu.cn).}
\thanks{Xiaopeng Fan is with the Faculty of Computing, Harbin Institute of Technology, Harbin 150001, P.R.China (e-mail: fxp@hit.edu.cn).}
\thanks{Jue Hu is with the National Key Laboratory of Science and Technology on Advanced Composites in Special Environments, Harbin Institute of Technology, Harbin 150080, P.R.China, and also with the Centre for Composite Materials and Structures, Harbin Institute of Technology, Harbin 150001, P.R.China (e-mail: juehundt@hit.edu.cn).}}

\markboth{Journal of \LaTeX\ Class Files,~Vol.~14, No.~8, August~2021}%
{Shell \MakeLowercase{\textit{et al.}}: A Sample Article Using IEEEtran.cls for IEEE Journals}

\IEEEpubid{0000--0000/00\$00.00~\copyright~2021 IEEE}

\maketitle

\begin{abstract}
Place recognition is a fundamental task for robotic application, allowing robots to perform loop closure detection within simultaneous localization and mapping (SLAM), and achieve re-localization on prior maps. Current range image-based networks use single-column convolution to maintain feature invariance to shifts in image columns caused by LiDAR viewpoint change. However, this raises the issues such as “restricted receptive fields” and “excessive focus on local regions”, degrading the performance of networks. To address the aforementioned issues, we propose a lightweight circular convolutional Transformer network denoted as CCTNet, which boosts performance by capturing structural information in point clouds and facilitating cross-dimensional interaction of spatial and channel information. Initially, a Circular Convolution Module (CCM) is introduced, expanding the network’s perceptual field while maintaining feature consistency across varying LiDAR perspectives. Then, a Range Transformer Module (RTM) is proposed, which enhances place recognition accuracy in scenarios with movable objects by employing a combination of channel and spatial attention mechanisms. Furthermore, we propose an Overlap-based loss function, transforming the place recognition task from a binary loop closure classification into a regression problem linked to the overlap between LiDAR frames. Through extensive experiments on the KITTI and Ford Campus datasets, CCTNet surpasses comparable methods, achieving Recall@1 of 0.924 and 0.965, and Recall@1\% of 0.990 and 0.993 on the test set, showcasing a superior performance. Results on the self-collected dataset further demonstrate the proposed method's potential for practical implementation in complex scenarios to handle movable objects, showing improved generalization in various datasets.
\end{abstract}


\begin{figure}[ht]
\centering
\includegraphics[width=3.4in]{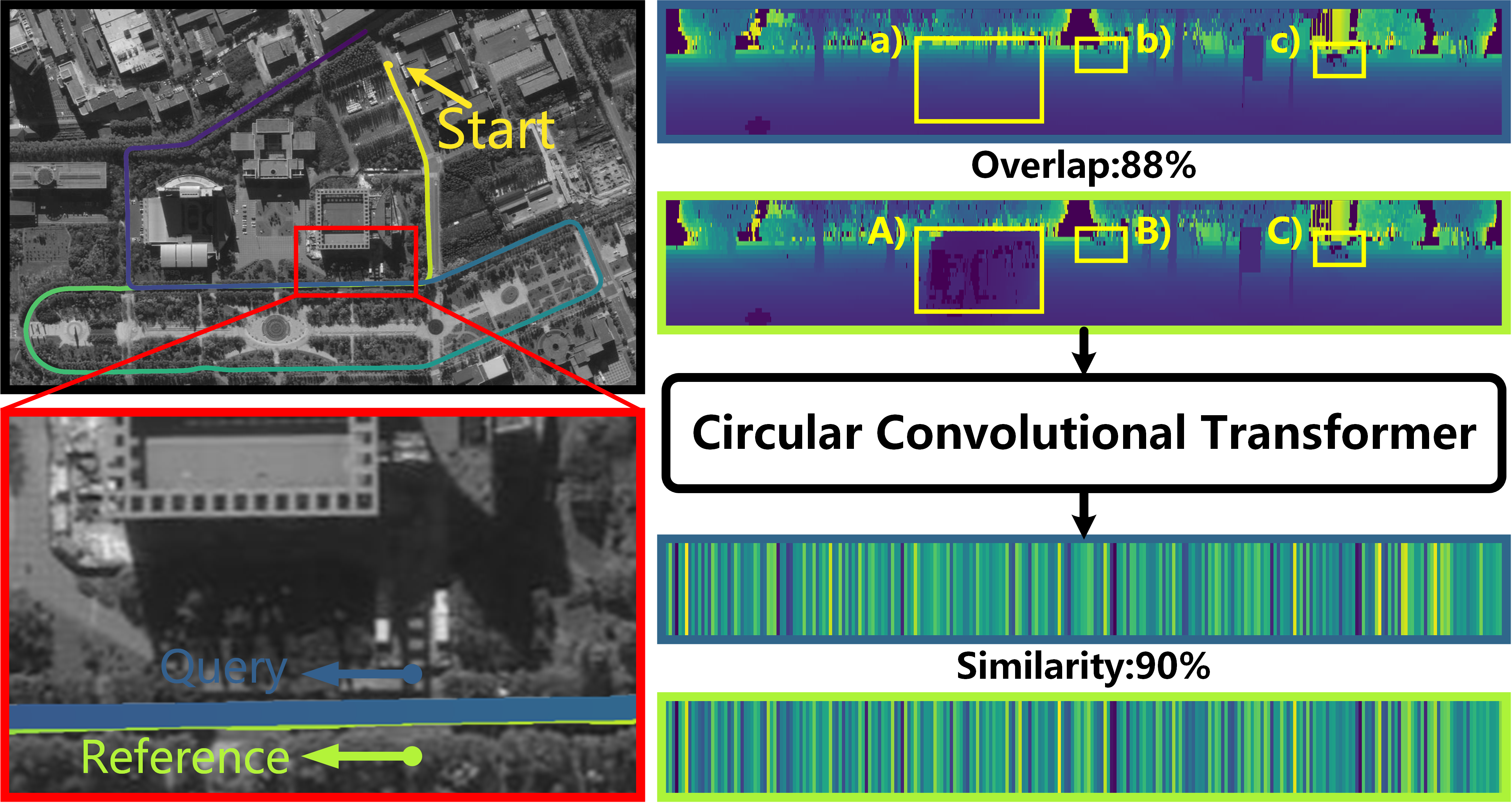}
\caption{Query frame (blue) and reference frame (green) at adjacent locations with movable objects. 1) Occlusions at A), B), and C) typically occupy multiple pixel columns. The limited receptive perception of single-column convolution hinders the network from capturing global information. 2) Despite the two frames originating from the same place, the presence of movable objects at A) compared to a) captured by a single spatial attention mechanism leads to higher weights allocated to that region. Our circular convolutional Transformer network is able to broaden the horizontal receptive field, and facilitate cross-dimensional interaction of spatial and channel information.}
\label{fig_1}
\end{figure}

\section{Introduction}
\IEEEPARstart{P}{lace} recognition refers to an intelligent robot determining if the current observed scenario has been visited before. This is accomplished by comparing the current sensor measurements, such as camera and light detection and ranging (LiDAR), with past measurements stored in a map or a database. It plays a crucial role in minimizing cumulative errors in simultaneous localization and mapping (SLAM) tasks \cite{RN1,RN2,RN3} and provides essential initial position guesses for global localization in Global Navigation Satellite System (GNSS)-denied conditions \cite{RN4,RN5}. Many vision-based place recognition methods \cite{RN6,RN7,RN8,RN9,RN10} have been proposed due to the widespread use of cameras. However, these methods face challenges in lighting conditions, as variations in illumination can impact feature extraction, leading to retrieval failures \cite{RN11, RN12}.

\IEEEpubidadjcol
In contrast to cameras, LiDAR draws substantial research interest due to its immunity to environmental lighting conditions, sparking extensive discussions on LiDAR-based place recognition. Existing research has thoroughly explored LiDAR-based place recognition methods relying on manual features \cite{RN11,RN13,RN14}, which are derived from researchers' in-depth understanding and analysis of sensing characteristics, e.g., the descriptors generated by techniques such as voxelization, histograms, mapping, and etc. For instance, Magnusson et al. \cite{RN13} classified point cloud cells into lines, planes, and spheres, which are subsequently aggregated into a feature vector to generate manual descriptors. Röhling et al. \cite{RN14} proposed a statistical-based method called Fast Histogram algorithm, which generates a one-dimensional histogram as a descriptor by directly counting the height information of the point cloud. Moreover, Scan Context \cite{RN11} employed the polar coordinate to map the point cloud into a two-dimensional (2D) matrix along radial and angular directions, serving as descriptors for place recognition. However, crafting manual features usually requires domain-specific expertise, and manual descriptors exhibit limited robustness in handling variations and uncertainties in complex scenes \cite{RN15}. They become vulnerable in scenarios involving movable objects, such as pedestrians or vehicles causing occlusion, or instances like parked vehicles changing positions.

Recent advances in deep learning have led to the application of relevant theories and methodologies for end-to-end descriptor generation. Many methods \cite{RN16,RN17,RN18} autonomously learn multi-level representations from point cloud data, capturing patterns of variations and occlusions in complex scenarios. In order to meet the real-time requirements, recent methods utilize range images generated from point cloud as inputs. OverlapNet \cite{RN19}, proposed by Chen et al., directly estimates the overlap and relative yaw angle between LiDAR frames for achieving place recognition. To address padding’s impact on the convolutional results when two point cloud frames undergo viewpoint changes, a single-column convolution kernel is used in this work. However, the issue of "restricted receptive field" limits capturing target-level information in the current region and hinders horizontal receptive field expansion through layer stacking. This limitation further hampers the extraction of global information for the current scene and disrupts structural relationships between columns of point clouds. Ma et al. \cite{RN20} further enhance the network’s learning capability for acquiring global features by combining OverlapNet with vision transformer (VIT) \cite{RN21}. They additionally incorporate NetVLAD \cite{RN22} in the network to ensure permutation invariance. Despite generating rotation-invariant descriptors, the network has an “excessive focus on local regions”. The pixel number of objects in the range image is inversely proportional to their distance from the sensor. As movable objects approach the sensor, their image proportion increases, reducing the background proportion. This variation causes spatial attention mechanisms to focus on specific image regions within the same place under occlusion, leading to diverse weights and compromising network performance. The two issues mentioned above are further illustrated in Fig. \ref{fig_1}.

\IEEEpubidadjcol
In addition, current research treats place recognition as a binary classification problem during training, categorizing the current location as either visited or not. Indeed, determining if the current place has been visited is linked to the distance between these places. Thus, the place recognition problem should be approached as a regression task reflecting the relationship between distance and similarity. Insofar as a deep network with higher accuracy and enhanced robustness against movable objects to be achieved, this work should wish to provide solutions to all three issues.

In this paper, a circular convolutional Transformer network with a regression loss is proposed for place recognition task in scenarios with movable object occlusion. Firstly, a Circular Convolution Module is proposed. It treats the range image as a ring, utilizing multi-column convolution to learn local feature details, relationships between range image columns, and circular structural features of the point clouds. Subsequently, a Range Transformer module is proposed to dynamically allocate weights to various channels and pixel regions, enabling the fusion and interaction of information from both channel and spatial dimensions. Finally, a regression loss function is introduced, using the overlap between two frames of point clouds as precise labels for the similarity of the generated descriptors. Extensive experiments are conducted on KITTI \cite{RN48}, Ford \cite{RN49}, and a self-collected dataset, validating the generalization capability of the proposed method.

The contribution of the proposed work can be summarized as three-fold:

a) A novel circular convolutional approach is proposed, which extends the horizontal receptive field while preserving point cloud structural features, addressing issues with traditional convolutions affected by viewpoint changes.

b) An interactive Transformer module is introduced across the spatial-channel dimensions, balancing global interactive representations between channels and spatial attention mechanism.

c) A loss function correlated with point cloud overlap is designed, eliminating the need to balance positive and negative samples and addressing the long-tailed effect.

The remaining structure of the paper is as follows: Section \uppercase\expandafter{\romannumeral2} further introduces relevant research in the field of place recognition, Section \uppercase\expandafter{\romannumeral3} presents the designed circular Transformer network in scenarios with occluded movable objects, and Section \uppercase\expandafter{\romannumeral4} provides experimental evaluations on the KITTI, Ford Campus, and a self-collected dataset (JLU Campus). Finally, the conclusion and summary are presented in Section \uppercase\expandafter{\romannumeral5}.

\section{Related Work}
In recent years, extensive research has focused on place recognition in autonomous driving. Some studies leveraged camera data, employing image processing and machine learning algorithms to discern surrounding scenes \cite{RN23,RN24,RN25,RN26,RN27,RN28}. Other studies, on the other hand, focused on using LiDAR \cite{RN11,RN19,RN20,RN29,RN30}, which provides three-dimensional (3D) coordinates for improving place recognition accuracy. Given its superior accuracy and stability over cameras in place recognition tasks, as well as its insensitivity to illumination and weather conditions, this paper primarily discusses methods based on 3D LiDAR point clouds.

\subsection{Place recognition based on manual features}
Methods based on manual features generally generate descriptors relying on attributes such as the distribution, shape, height and intensity of point clouds \cite{RN11,RN29,RN30,RN31}. These descriptors are then utilized to retrieve corresponding place information from a pre-built map or database. Kim et al. \cite{RN11} introduced Scan Context (SC), which projects 3D point clouds onto a 2D grid, with the value of each grid determined by its maximum height. Kim et al. \cite{RN29} further extended SC with Scan Context++ (SC++), which projects point clouds into polar or Cartesian coordinates based on the scenario-polar for viewpoint changes and Cartesian for lane changes. The existing methods mentioned above concentrate solely on the maximum height of point clouds, neglecting the information of LiDAR intensity. Therefore, Wang et al. \cite{RN31} proposed Intensity Scan Context (ISC), which incorporates LiDAR intensity into SC and proposes a two-stage hierarchical intensity retrieval strategy. This strategy utilizes fast binary operations to expedite the place recognition process.

As the aforementioned methods require brute-force matching during the retrieval phase, it compromises the efficiency and accuracy of scene recognition. Wang et al. \cite{RN30} proposed LiDAR Iris, which avoids brute-force matching through Fourier transformation and ensures rotation invariance. They further used a LoG-Gabor filter to generate a binary feature map for each frame, achieving similarity calculation between descriptors. With the continuous advancement in hardware, Rublee et al. \cite{RN32} utilized a high-resolution imaging LiDAR (VLS-128) to obtain images based on LiDAR intensity. They treated high-resolution point clouds with intensity as images and utilized visual methods to achieve place retrieval. However, this method imposes high hardware requirements, where the point cloud resolution significantly influences both the quality of images and the performance of method. Gao et al. \cite{RN33} introduced FP-Loc, which relies on robust ceiling and floor detection. It can address partial pose estimation and support segmentation of vertical structures, such as walls and columns. Unfortunately, it can not be used for outdoor localization due to its reliance on planes in scenarios.

\subsection{Place recognition based on deep learning}
\textbf{1) Point cloud based deep learning models.} 
As deep learning advances in image processing, its application is also extended to the 3D domain \cite{RN34,RN35,RN36,RN37}. Uy et al. \cite{RN12} proposed PointNetVLAD, which is the first to employ deep learning for place recognition. It employs PointNet for features extraction from point clouds and encodes them with NetVLAD to obtain the global descriptor. To enhance the feature extraction capacity of PointNet, Komorowski et al. \cite{RN38} proposed MinkLoc3D. It utilizes sparse 3D convolution for feature extraction from the sparse point cloud representation, and uses a Feature Pyramid Network (FPN) \cite{RN38} architecture for descriptor generation. MinkLoc3D-SI, proposed by Żywanowski et al. \cite{RN39}, employs a spherical representation for point clouds and integrates intensity to improve single-shot place recognition. Inspired by the U-Net, Vidanapathirana et al. \cite{RN40} proposed LoGG3D-Net, an end-to-end network that encodes point clouds features and generates descriptors through 3D sparse convolution.

Leveraging the successful applications of Transformer model in image processing, particularly ViT \cite{RN21}, Swin \cite{RN41}, Zhou et al. \cite{RN18} designed a Transformer-based network to learn global descriptors with contextual information via Normal Distributions Transform (NDT) cell generated from point clouds. Hou et al. \cite{RN42} proposed a Hierarchical Transformer for place recognition, dividing point clouds into small units through downsampling and nearest neighbor searching. The proposed structure was then used to extract local and global features within and between these units.

\textbf{2) Range image based deep learning models.}
Using point clouds as direct input for deep neural network models poses a challenge due to high computational complexity, limiting real-time place recognition. Therefore, Chen et al. \cite{RN19} proposed OverlapNet, which employs a 2D convolutional neural network calculating the overlap and yaw between two scans. The model takes the multi-channel image as input, generated from point cloud depth, normals, intensity, and semantic information. To achieve high-precision global localization based on LiDAR sensors, Chen et al. \cite{RN43} futher utilized OverlapNet as the observation model for the Monte Carlo Localization (MCL) algorithm. The network predicts the overlap and yaw between two scans of point clouds to update particle weights, thereby improving the performance of MCL. However, the processing of semantic information may not meet real-time standards. To tackle this, Ma et al. \cite{RN20} proposed OverlapTransformer, using range images as the sole input to achieve fast and real-time place recognition. They employed an enhanced OverlapNet for range image encoding, followed by feeding the results into a Transformer module based on Vision Transformer (ViT). The descriptor vector is then generated using the NetVLAD structure. Ma et al. \cite{RN44} proposed CVTNet, to fuse the range image views (RIVs) and bird’s eye views (BEVs) generated from the LiDAR data. It extracts correlations within the views using intra-transformers and between the two different views.

\textbf{3) Semantic information based deep learning models. }
As semantic segmentation tasks with point clouds advance, numerous methods leveraging semantic information for place recognition have emerged.  Li et al. \cite{RN45} proposed Semantic Scan Context (SSC), which encodes advanced semantic features through kd-tree, effectively leveraging semantic information between point clouds to represent 3D scenarios. Cramariuc et al. \cite{RN46} introduced SemSegMap, a method inferring point cloud semantics from image semantic information. It then utilized PointNet++ for feature extraction and descriptor generation. Object Scan Context (OSC), proposed by Yuan et al. \cite{RN47}, involved  semantic segmentation on point clouds to identify objects (e.g., cylindrical objects). Descriptors were then constructed via these objects, ensuring independence from observation positions for accuracy improvement of place recognition. However, acquiring semantic information requires extra computations. In the case of autonomous vehicles with limited onboard computing resources, ensuring real-time performance becomes challenging. In contrast, our approach relies solely on raw range information to achieve real-time performance, making our method more easily applicable to diverse datasets collected by various LiDAR sensors.

\begin{figure*}[ht]
\centering
\includegraphics[width=6.7in]{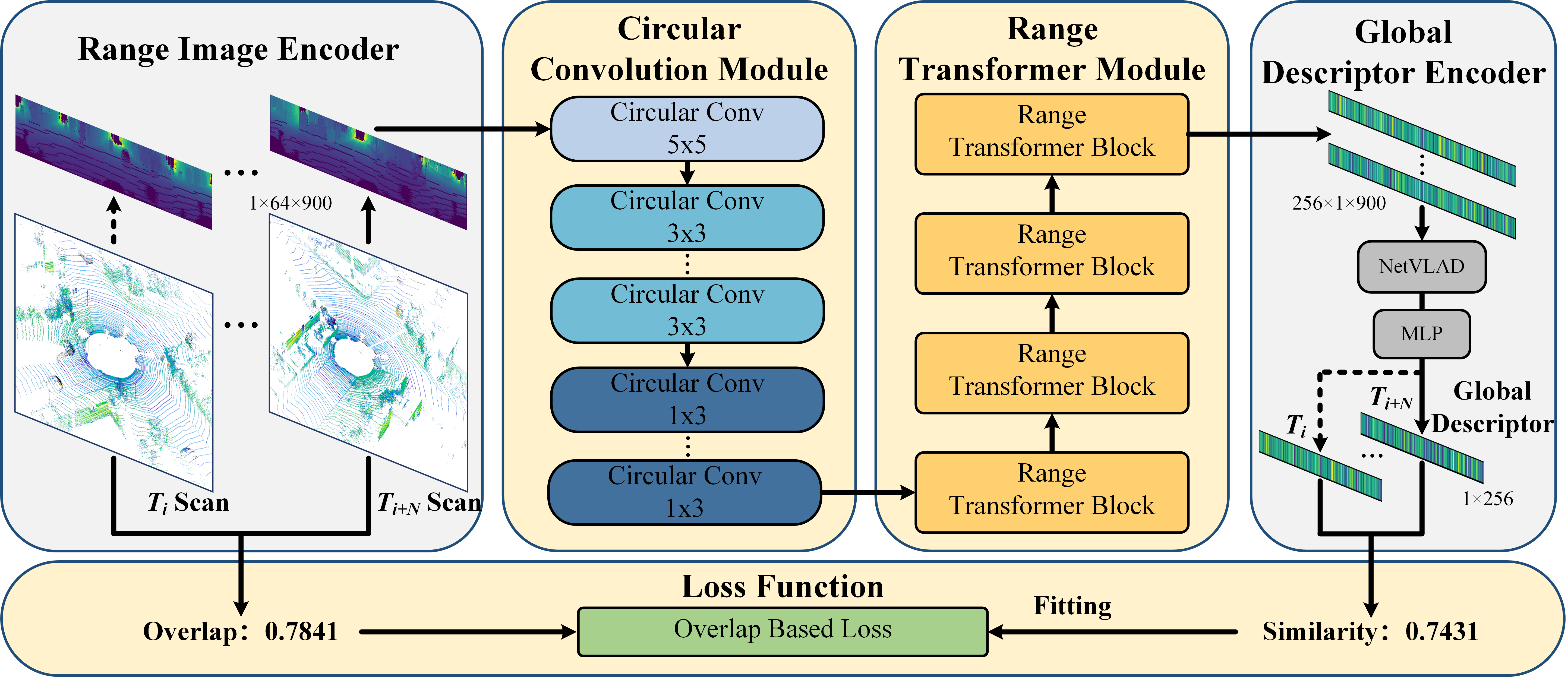}
\caption{The pipeline of the proposed method. The highlighted orange section comprises the main content of this paper.}
\label{fig_2}
\end{figure*}

\section{Method}
In this section, a Transformer-based localization architecture is proposed, including five stages: 1) Range Image Encoder, 2) Circular Convolution Module, 3) Range Transformer Module, 4) Global Descriptor Encoder, and 5) Loss Function. In the Range Image Encoder stage, the point cloud is converted into a yaw-angle-equivalent range image. Next, the Circular Convolution Module treats the range image as a 360° panoramic image, extracting the circular structural features and column relationships. Then, the Range Transformer Module, receiving the output from the Circular Convolution Module, mitigates the influence of movable objects on the feature map using channel and spatial attention mechanisms. The feature map is further sent to the Global Descriptor Encoder, which utilizes NetVLAD and MLP to transform it into a global descriptor of the scene. During the training process, the overlap between point clouds serves as the label for generated descriptor similarity, enhancing the model's domain adaptability. Fig. \ref{fig_2} illustrates the overall pipeline of the proposed method. The detailed implementation of the proposed method are elaborated below.

\subsection{Range Image Encoder}
\label{section:A}
To reduce computation, it is necessary to transform the point clouds $P$ into range images $R$. The mapping progress can be divided into two steps. Each point $p(x,y,z)$ needs to calculate its mapped coordinate $(u,v)$ in the range image during the first step. The pixels at each coordinate are then assigned their values in the second step. Then we have:
\begin{equation}
\begin{split}
    \label{eq1}
    \left( u,v \right)=\big( \frac{w}{2} \cdot \left[ 1-\frac{1}{\pi} \cdot {\arctan (\frac{y}{x})} \right], \\ h \cdot \left[ 1-\frac{1}{f} \cdot (\arcsin (\frac{z}{\left\| {p} \right\|})+{{f}_{up}}) \right] \big),
\end{split}
\end{equation}
\begin{equation}
\label{eq2}
R(u,v)=\left\| {p} \right\|,{p}\in P,
\end{equation}
\noindent
where $w$ and $h$ represent the width and height of the range image respectively, $u\in [1,w]$, $v\in [1,h]$. ${f}={{f}_{up}}+{{f}_{down}}$ is the vertical field of view of the LiDAR, ${{f}_{up}}$ is the maximum elevation angle of the LiDAR's field of view, while ${{f}_{down}}$ is the maximum depression angle of the LiDAR's field of view. 

It is worth noting that, the mapping progress is performed along the angular direction. Hence, the rotation between point clouds manifests as a column shift between range images, which is yaw-angle equivariant.

\subsection{Circular Convolution Module}\label{3b}
Mechanical LiDAR achieves a 360° horizontal field of view through continuous rotation. Theoretically, the range image generated from point clouds should form a panoramic view with both ends of the image are continuous \cite{RN52}. However, existing methods employ single-column convolution for feature extraction from the range image, which neglects the inter-column relationships due to the small receptive field. They struggle to learn the correlation between pixels within the current convolution kernel and the entire image. In addition, they fail to learn ring structure features in the point cloud due to the disconnection at both ends of the image.

\begin{figure*}[ht]
\centering
\includegraphics[width=6.8in]{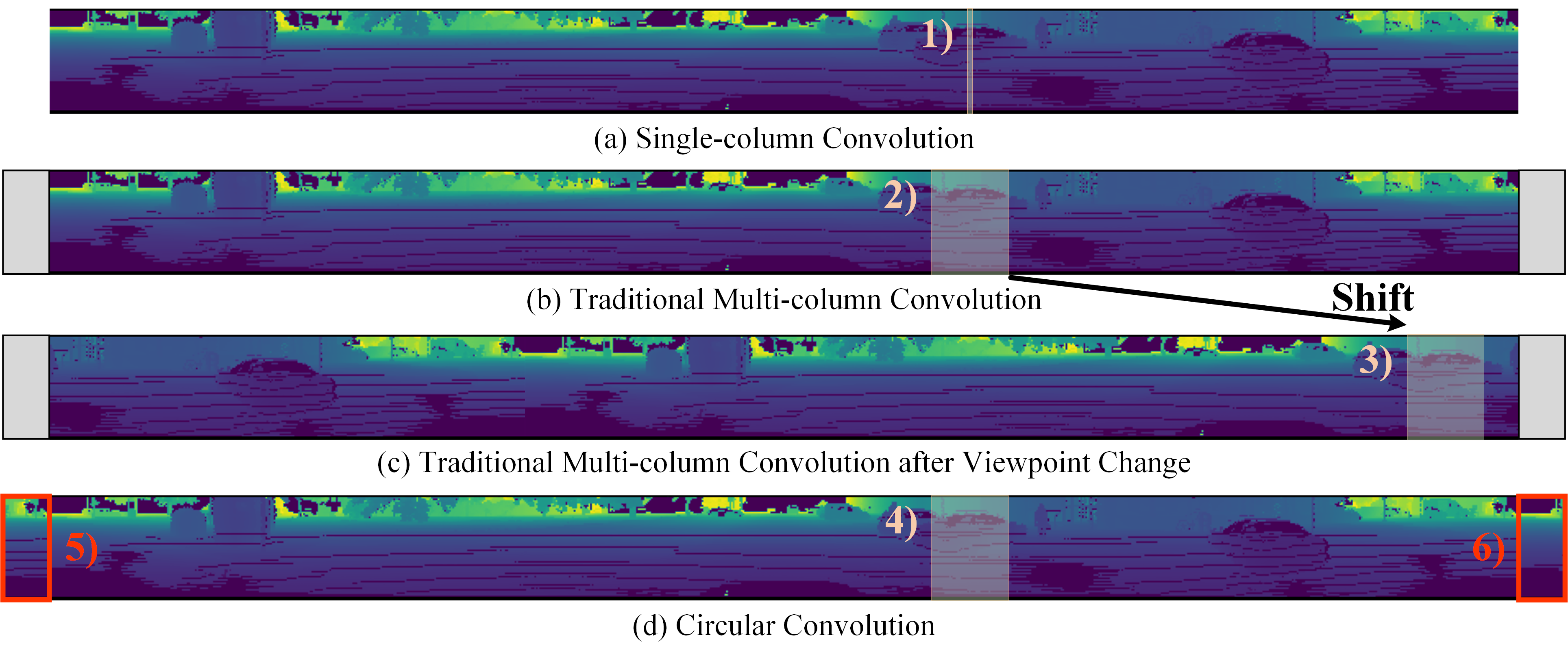}
\caption{Illustration of single-column convolution, traditional multi-column convolution, and circular convolution. (a) Given the small receptive field, it is challenging to capture contextual information between the kernel and the entire image. In (b) and (c), the viewpoint change of LiDAR causes column shifts in the range image, leading to changes introduced by zero-padding at both ends during feature extraction. (d) Circular convolution transforms the range image into a 360° panoramic image, which effectively addresses issues related to viewpoint changes, ensuring consistent features extracted from multiple columns.}\vspace{-5mm}
\label{fig_3}
\end{figure*}

Fig. \ref{fig_3} (a) demonstrates a convolution kernel with a column width ${k_w=1}$. To capture relationships between image columns, enlarging the column width $k_w$ of the convolution kernel is essential, as indicated by the rectangle labeled 2) in Fig. \ref{fig_3} (b). In traditional convolutional methods, zero-padding is applied to both ends to avoid width shrinkage in the convolved image, depicted by the gray areas in Fig. \ref{fig_3} (b). However, when the LiDAR viewpoint changes, the range image undergoes column shift, as seen in 2) and 3) in Fig. \ref{fig_3} (b) and \ref{fig_3} (c). In this case, adding zero-padding at both ends of Fig. \ref{fig_3} (b) and (c) leads to varying padding positions for images generated from the same place under different viewpoints. This further leads to different convolution results, which is unacceptable for place recognition.

\begin{figure}[h]
\centering
\includegraphics[width=3.4in]{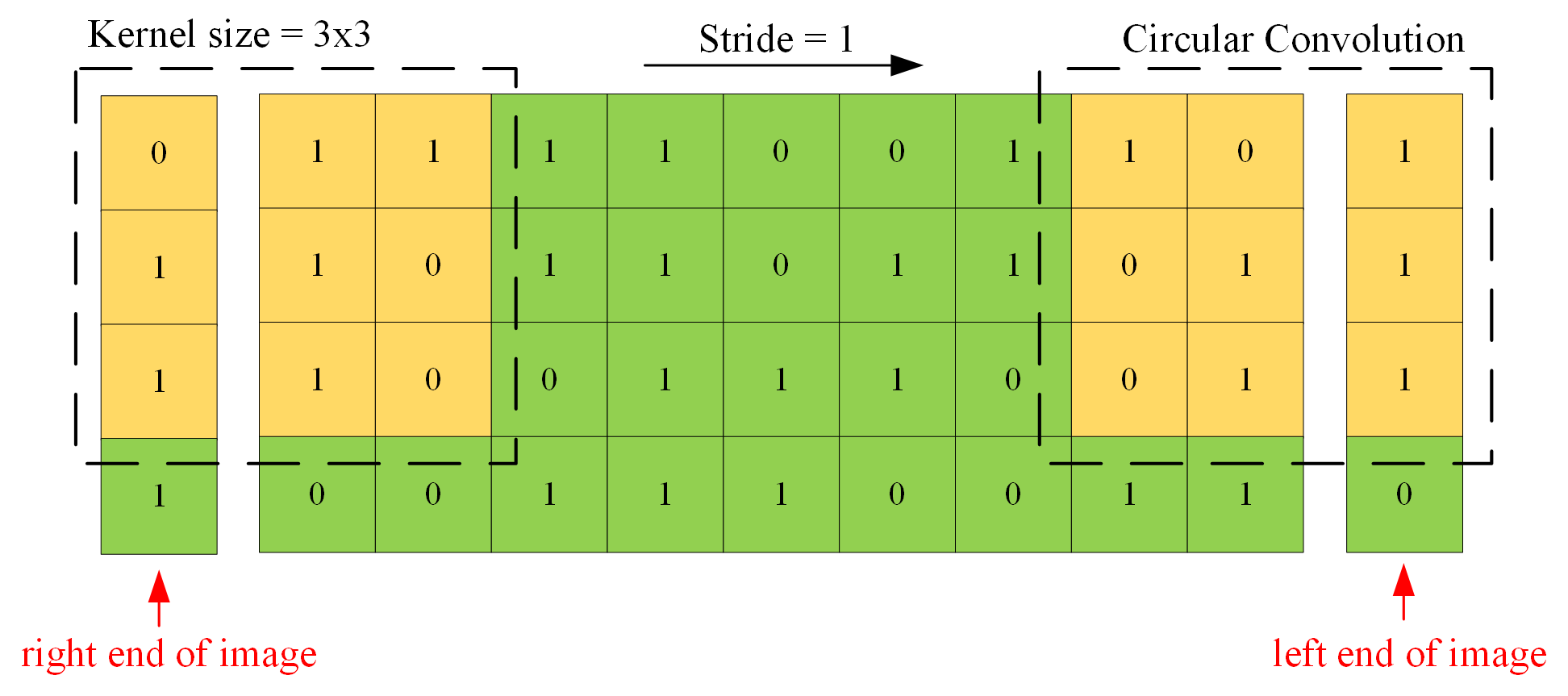}
\caption{Illustration of circular convolution. The circular range image is first generated by expanding both ends of the range image, followed by performing convolution using kernels with a fixed stride.}
\label{fig_4}
\end{figure}

Hence, the Circular Convolution Module $F_{CCM}$ is introduced in this paper, depicted by the red boxes labeled 5) and 6) in Fig. \ref{fig_3} (d). This module adaptively pads one side of the image to the other based on the convolution kernel size and stride, allowing the kernel to span both ends of the image. This ensures that the convolution results remain unaffected by changes in viewpoint. The $F_{CCM}$ takes the circular range image $R_{circ}$ as input and outputs the feature map $A_{CCM}$, then we have:
\begin{equation}
\label{eq3}
{{A}_{CCM}}={{F}_{CCM}}({{R}_{circ}}),
\end{equation}
\noindent
where $R_{circ}$ is a ring-shaped range image extended from range image $R$, $R_{circ}=[{R_{:,(w-{{pad}_{left}}):(w-1)}\ R\ R_{:,0:({pad}_{right}-1)}}]$. $w$ is the width of the range image. ${pad}_{left}$ and ${pad}_{right}$ represent the number of padded columns on the left and right sides of the image, respectively, which are given by:
\begin{align}
\label{eq45}
   & {{pad}_{left}}=\left\lfloor {{{pad}_{w}}}/{2}\; \right\rfloor, \\ 
   & {{pad}_{right}}={{pad}_{w}}-{{pad}_{left}},
\end{align}
\noindent
${{pad}_{w}}$ denotes the total padding width, yielding the following padding rules:
\begin{equation}
\label{eq6}
{{pad}_{w}}=\begin{cases}
  \max({{K}_{w}}-{{S}_{w}},0),\text{ } \mathbf{if }\text{ } w\text{ }\bmod \text{ }{{S}_{w}}=0 \\  \max({{K}_{w}}-w\text{ }\bmod \text{ }{{S}_{w}},0),\text{ } \mathbf{otherwise}
\end{cases},
\end{equation}
\noindent
where $K_w$ is the width of the convolutional kernel, and $S_w$ is the horizontal stride. The specific implementation of circular convolution is described in Fig. \ref{fig_4}, and parameters of different convolution kernels in the Circular Convolution Module are recorded in Table \ref{tb_1}.

\begin{table}[h]
\centering
\caption{Parameters of convolution kernels in the Circular Convolution Module.}
\label{tb_1}
\begin{tabular}{cccc}
\hline
Name of the kernel & Occurrences in CCM & Padding size & Stride \\ \hline
5×5                & 1                  & 2×2          & 2×1    \\
3×3                & 5                  & 1×1          & 2×1    \\
1×3                & 3                  & 1×0          & 1×1    \\ \hline
\end{tabular}
\end{table}

\subsection{Range Transformer Module}
Objects in the scenario contribute differently to place recognition. Traditional convolution excels at extracting local features from images but may struggle to prioritize essential regions. To overcome this limitation, current methods generally leverage the ViT \cite{RN21} network, known for its spatial attention capabilities. ViT \cite{RN21} assigns different weights to pixels in distinct regions of the image, dividing the image into multiple regions with varying importance levels. However, spatial attention emphasizes pixels in specific regions, causing feature differences in scenes with or without movable objects and resulting in descriptor variations. To alleviate this issue, a Range Transformer Module, combining channel attention and spatial attention mechanisms, is introduced. The channel attention dynamically allocates weights to features, mitigating the local focus of spatial attention. This contributes to consistent descriptor generation.

\begin{figure*}[ht]
\centering
\includegraphics[width=6in]{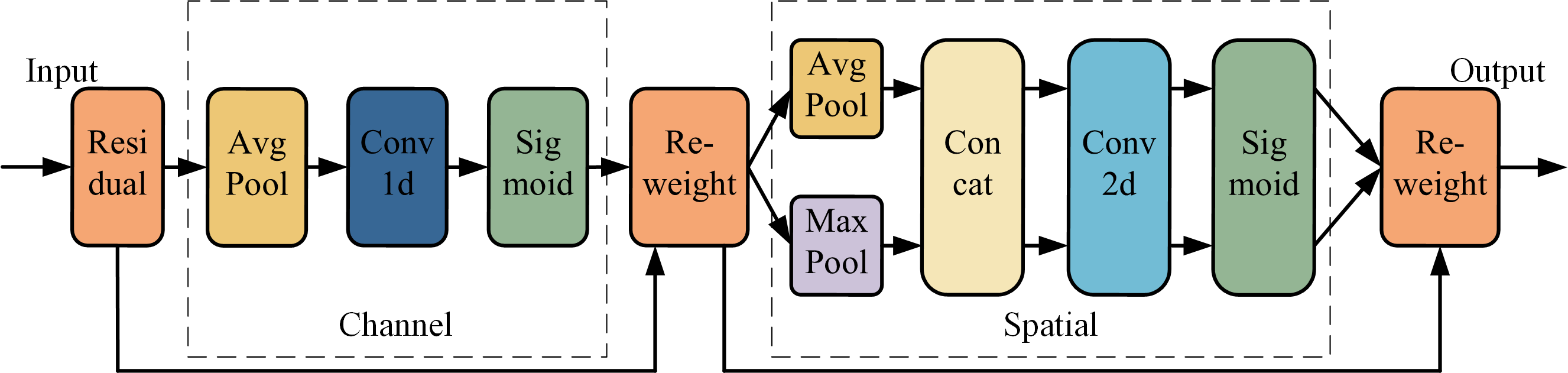}
\caption{RTM attention structure schematic. The channel attention of RTM consists of average pooling, 1D convolution, and a Sigmoid activation function. The spatial attention concatenates the outputs of average pooling and max pooling along the channel dimension. After 2D convolution and Sigmoid activation, the attention weight tensor is calculated, which is then used to reassign the feature map weights.}\vspace{-5mm}
\label{fig_5}
\end{figure*}

Specifically, as depicted in Fig. \ref{fig_5}, the feature map $A_{CCM}$ extracted from the upstream CCM is then fed into a cascaded attention mechanism and we have:
\begin{equation}
\label{eq7}
{{A}_{RTM}}={{F}_{RTM}}({{A}_{CCM}})\otimes {{A}_{CCM}},
\end{equation}
\noindent
where $\otimes$ denotes the re-weight operation of the feature map.

Although local features are extracted from the distance image, yaw-invariant descriptors are not generated. The generation of a yaw-angle equivariant feature is a prerequisite for creating yaw-angle invariant descriptors. In Sec. \ref{3b}, we discussed that CCM can extracts yaw-angle equivariant features, ensuring consistent feature values. Similarly, RTM, by dividing the feature map into patches and using cascaded attention mechanisms, can also generate these features. However, they can not preserve consistent column order in the case of LiDAR viewpoint rotation. To address this issue, we further feed features $A_{RTM}$ into NetVLAD \cite{RN22} and Multi-Layer Perceptron (MLP) to produce a rotation invariant one-dimensional descriptor $D$. The following equation illustrates this process:
\begin{equation}
\label{eq8}
D={F_\text{MLP}}({F_\text{NetVLAD}}({{A}_{RTM}})).
\end{equation}
\subsection{Overlap-Based Loss Function}
Range image based methods generally supervise the network using the overlap, which is a more natural way to describe the similarity between two LiDAR scans compared to distance \cite{RN19,RN20}.  However, these networks treating place recognition as a binary classification problem. The overlap between scan pairs surpasses a certain threshold, they are considered as positive samples, vice versa. Indeed, the positions corresponding to positive samples are physically closer, indicating a higher degree of overlap between them. Employing continuous labels within the range of 0 to 1 is more appropriate than utilizing discrete binary values of 0 and 1. Hence, treating place recognition as a regression problem aligns better with our practical understanding, where the probability of loop closure between pairs of point clouds correlates with their overlap. Based on this, we propose an overlap-based regression loss function.

Let the query image $R_q^i$ and the reference image $R_r^j$ be a pair in the training set. $D_q^i$ and $D_r^j$, are descriptors generated by the proposed network from $R_q^i$ and $R_r^j$, respectively. Then, the loss function can be given in a regression manner as: 
\begin{equation}
\label{eq9}
loss=\sum{_i\sum{_j\left\|{{Sim({{D}^{i}_{q}},{{D}^{j}_{r}})-Over({{R}^{i}_{q}},{{R}^{j}_{r}})}}\right\|}}.
\end{equation}
Eq. (\ref{eq9}) shows the proposed loss function consists of two terms: one term measures overlap, the other measures similarity. The $Over(\cdot)$ function follows the approach described in OverlapNet \cite{RN19}. It determines the overlap by comparing the query image $R_q$ with the reprojected image $R_{r}^{’}$, obtained by projecting the reference range image $R_r$ into the query image coordinate:
\begin{small}
\begin{equation}
\label{eq10}
Over({{R}_{q}},{{R}_{r}})=\frac{\sum{_{(u,v)}\Omega \left( \left\| {{R}_{q}}(u,v)-{{{{R}'}}_{r}}(u,v) \right\|\le \delta  \right)}}{\min (valid({{R}_{q}}),valid({{{{R}'}}_{r}}))},
\end{equation}
\end{small}
\noindent
where  $\Omega(a)=1$ if $a$ is true and $\Omega(a)=0$ otherwise. $valid(\cdot)$ calculates the number of valid pixels (values grater than 0) in the range image. $\delta$ serves as a predefined threshold for deciding overlapped pixels.

The similarity between descriptors is calculated through $Sim(\cdot)$ function, which is given as:
\begin{equation}
\label{eq11}
Sim({{D}_{q}},{{D}_{r}})=\frac{1}{2}\left({\frac{{{{D}_{q}}}\cdot {{{D}_{r}}}}{| {{{D}_{q}}} |\cdot | {{{D}_{r}}} |}+1}\right).
\end{equation}
\subsection{Testing}
After training, the network model parameters converge, making the model ready for testing. Given a set of point clouds representing places previously visited by the vehicle. They are subsequently transformed into range images and further fed into the trained network to generate a set of descriptors $\mathcal{D}$. For any query point cloud, originating from the current place of vehicle, it first undergoes Range Image Encoder to generate $R_q$, then feeds into the network to produce descriptor $D_q$. To identify the data in the point cloud set that is most likely to come from the same place as the query point cloud, the place is determined by finding the descriptor $\hat{D}$ in $\mathcal{D}$ that has the highest similarity with $D_q$, derived the following equation:
\begin{equation}
\label{eq12}
\hat{D}=arg\max_{D_{i}\in{\mathcal{D}}}{(Sim(D_q,D_i))}.
\end{equation}
\section{Experiments}

This section evaluates the performance of the proposed method through extensive experiments on the KITTI \cite{RN48}, Ford Campus \cite{RN49}, and a self-collected dataset (JLU Campus). The proposed method is compared with the state-of-the-arts, and the ablation study is also performed to demonstrate the improvement of each component of the proposed method.

\subsection{Datasets and Experimental Setup}
\textbf{1) Datasets:}

KITTI dataset \cite{RN48}: it is a well-known public benchmark dataset, which is collected in Karlsruhe, including scenarios like city streets, highways, and rural roads. Their data collection platform is equipped with a Velodyne HDL-64E rotating 3D laser scanner and an OXTS RT3003 inertial navigation system (INS), providing point clouds and accurate ground truth.

Ford Campus dataset \cite{RN49}: it is also a public dataset collected at the Ford Research Campus and downtown areas, using a modified Ford F-250 pickup truck. The vehicle provides point clouds generated by Velodyne-HDL64e LiDAR and ground truth provided by Applanix POS LV and Xsens MTI-G.

\begin{figure}[h]
\centering
\includegraphics[width=3in]{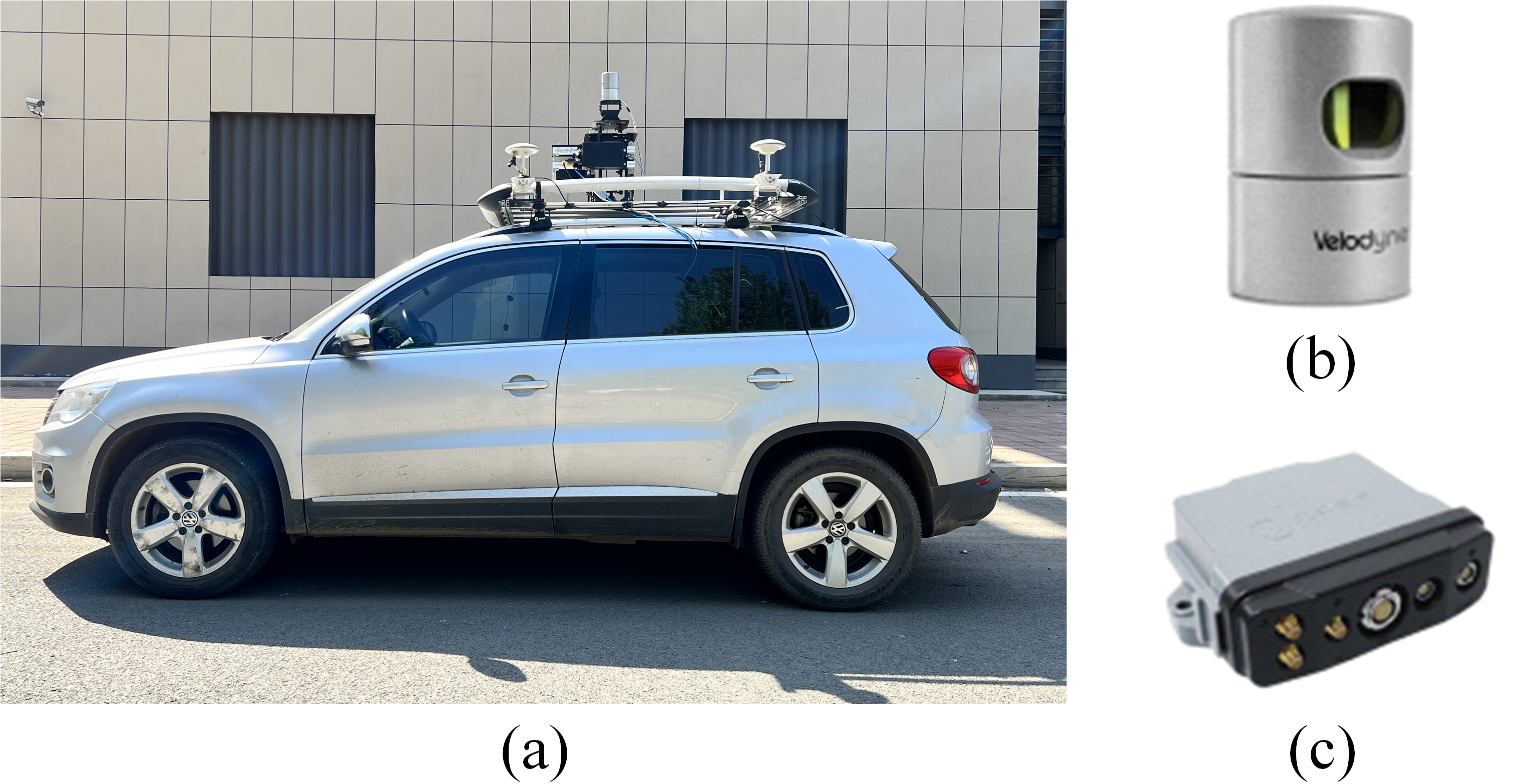}
\caption{Illustration of data collection platform: (a) the modified Volkswagen Tiguan vehicle, (b) Velodyne HDL-32E, and (c) Npos220s.}
\label{fig_6}
\end{figure}

JLU Campus dataset: it is a self-collected dataset created at Qianwei Campus of Jilin University, using a modified Volkswagen Tiguan vehicle (as illustrated in Fig. \ref{fig_6}). The platform generates point clouds by a Velodyne HDL-32e LiDAR and provides ground truth by a Npos220s.

\textbf{2) Environment:}

All experiments were conducted on a computer equipped with an ASUS Z9PA-D8 WS mainboard, dual Intel Xeon E5-2680 v2 @2.80 GHz processors, 64 GB of RAM, and a NVIDIA RTX 2080 graphics card. All codes are implemented in Python with PyTorch.

\textbf{3) Implementation Details:}

For the KITTI and Ford Campus datasets, the point cloud from a 64-beam LiDAR, is processed into range images with a size of 1$\times$64$\times$900. The JLU Campus dataset, which used a 32-beam LiDAR, processes the data into images with a size of 1$\times$32$\times$900. During the training process, the batch size is set to 12 (with 6 for $R_q$ and 6 for $R_r$). We employ the Adaptive Moment Estimation optimizer to dynamically adjust the learning rate of the parameters, with an initial learning rate of 1e-3.  For NetVLAD, the output feature dimension is set to 256, which allows our network to generate a global descriptor with same size. Regarding the threshold $\delta$ in Eq. (\ref{eq10}), it is set to 1 referring to the settings of OverlapTransformer \cite{RN20}.

\subsection{Evaluation on KITTI Dataset}
To quantitatively validate the performance of our method, we compare it with state-of-the-art open-source methods, including Fast Histogram \cite{RN14}, Scan Context \cite{RN11}, LiDAR Iris \cite{RN30}, PointNetVLAD \cite{RN12}, OverlapNet \cite{RN19}, NDT-Transformer \cite{RN18}, MinkLoc3D \cite{RN16}, OverlapTransformer \cite{RN20}, and CVTNet \cite{RN44}. All these methods are first trained and evaluated on KITTI dataset. Sequences 03-10 are employed for training, sequence 02 for validation, and sequence 00 for evaluation. Two scans will be regarded as a loop closure if their predicted overlap value is larger than 0.3. In addition, Recall@1 and Recall@1\% are utilized for evaluation metrics.

The results of our method and other methods are shown in Table \ref{Table2:KITTIAndFord}. As can be seen, the proposed method in this paper outperforms the compared methods. Our method achieves a Recall@1 result of 0.924, demonstrating a 1.43\% and 1.99\% improvement over state-of-the-art deep learning methods CVTNet and OverlapTransformer, respectively. Compared to the handcrafted feature methods Iris and Scan Context, our method exhibits an improvement of 10.66\% and 12.68\%, respectively. It indicates that when confronted with the constraint of selecting only the top-1 result, our method outperforms these traditional methods by correctly identifying 10.66\% and 12.68\% more instances. As for the metric of Recall@1\%, our method still outperforms other methods, with a value of 0.990. Surpassing the second best method CVTNet by 1.23\%, this means that when constrained to selecting only the top-ranked result among every hundred locations, our method outperforms CVTNet in correctly identifying locations, exceeding by 1.23 locations. The success can be attributed to the efficient feature extraction and stable descriptor generation achieved through the multi-column convolution strategy and the integration of channel and spatial attention mechanisms. This is especially beneficial in scenarios like the KITTI 00 dataset, where loop closure samples span significant intervals, leading to a higher probability of position changes in movable objects within the scene.

\begin{table}[]
\centering
\caption{Comparison of place recognition performance on Sequence 00 of Ford Campus Dataset and KITTI Dataset.}
\label{Table2:KITTIAndFord}
\begin{tabular}{@{}cccc@{}}
\toprule
Dataset & Approach              & Recall@1 & Recall@1\% \\ 
\midrule
\multirow{10}{*}{KITTI}                                                 
        & Fast Histogram \cite{RN14}    & 0.738    & 0.871 \\
        & Scan Context \cite{RN11}      & 0.820    & 0.869     \\
        & LiDAR Iris \cite{RN30}      & 0.835    & 0.877     \\
        & PointNetVLAD \cite{RN12}      & 0.776    & 0.845     \\
        & OverlapNet \cite{RN19}        & 0.816    & 0.908     \\
        & NDT-Transformer-P \cite{RN18}  & 0.802    & 0.869     \\
        & MinkLoc3D \cite{RN16}          & 0.876    & 0.920     \\
        & OT \cite{RN20} & 0.906    & 0.964     \\
        & CVTNet \cite{RN44}            & 0.911    & 0.978     \\
        & Ours                  & \textbf{0.924}    & \textbf{0.990}     \\ \midrule
\multirow{10}{*}{\begin{tabular}[c]{@{}c@{}}Ford\\ Campus\end{tabular}} 
        & Fast Histogram \cite{RN14}    & 0.812    & 0.897 \\
        & Scan Context \cite{RN11}      & 0.878    & 0.958     \\
        & LiDAR Iris \cite{RN30}        & 0.849    & 0.937     \\
        & PointNetVLAD \cite{RN12}      & 0.862    & 0.938     \\
        & OverlapNet \cite{RN19}         & 0.857    & 0.932     \\
        & NDT-Transformer-P \cite{RN18}  & 0.900    & 0.927     \\
        & MinkLoc3D \cite{RN16}  & 0.878    & 0.942     \\
        & OT \cite{RN20} & 0.914    & 0.954     \\
        & CVTNet \cite{RN44}    & 0.920    & 0.981     \\
        & Ours                  & \textbf{0.965}    & \textbf{0.993}     \\ \bottomrule
\end{tabular}
\end{table}

\subsection{Evaluation on Ford Campus Dataset}
To evaluate the performance of the proposed method in unseen environments and demonstrate its generation ability, all the model trained on KITTI dataset are directly tested on the Ford Campus dataset without fine-tuning. The results of the comparative methods and ours are detailed in Table \ref{Table2:KITTIAndFord}.

From Table \ref{Table2:KITTIAndFord}, it is evident that our method excels on the Ford Campus dataset, achieving the highest values for Recall@1 and Recall@1\%, namely 0.965 and 0.993, respectively. Compared to the traditional algorithm SC, our Recall@1 and Recall@1\% values show improvements of 9.91\% and 3.65\%, respectively. Moreover, compared to the deep learning algorithm CVTNet, the improvements are 4.89\% and 1.22\%, respectively. The primary factor contributing to the improvements is the Overlap-based loss function. This loss guides the generation of descriptor similarity by leveraging the Overlap value of two LiDAR scans, playing a crucial role in enhancing the model's domain adaptation capability.

\subsection{Evaluation on JLU Campus Dataset}
\begin{figure}[h]
\centering
\includegraphics[width=3.4in]{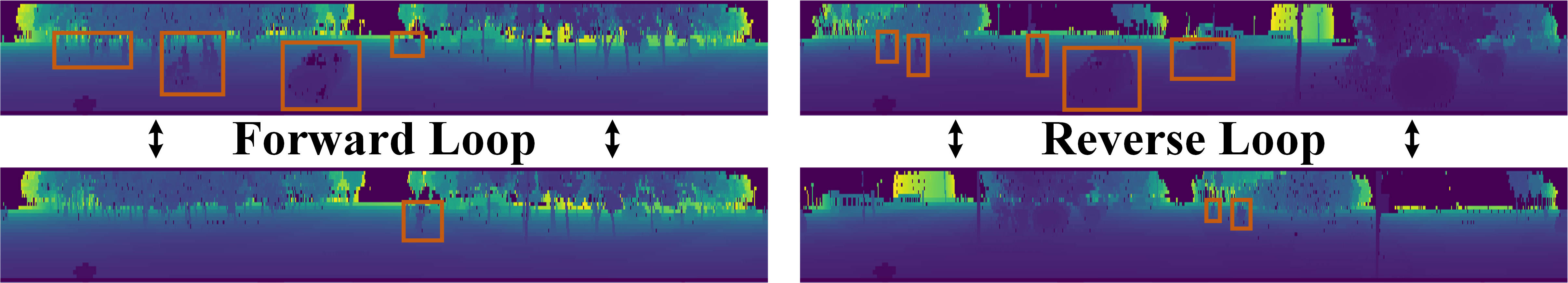}
\caption{Illustration of both both positive and negative loop closures in the JLU Campus Dataset. The brown boxes highlight movable objects.}
\label{fig_7}
\end{figure}

To further validate the proposed method can generalize well to the diverse environments with different LiDAR sensors, we conduct experiments using a self-collected dataset. For the JLU Campus dataset, sequence 03 has the highest number of loop closures, particularly featuring a substantial amount of reverse loop closures, making it an ideal dataset for model training. Sequence 02 is served as the validation set. Sequence 01 and 04 are designated for testing due to the presence of both forward and reverse loop closures. The criterion for setting ground truth is that the distance between the initial frame and the loop frame is less than 4 meters. The detailed information of these sequences is shown in Table \ref{table3}. The visualization in Fig. \ref{fig_7} provides additional insights into the dataset.

\begin{table*}[ht]
\centering
\caption{Comparison of place recognition performance on Sequence 01 and 04 of the JLU Campus Dataset.}
\label{Table4}
\resizebox{\textwidth}{!}{%
\begin{tabular}{@{}ccccccccccccccc@{}}
\toprule
\multirow{2}{*}{Approach} & \multicolumn{7}{c}{Sequence 01}                              & \multicolumn{7}{c}{Sequence 04}                              \\ \cmidrule(l){2-8} \cmidrule(l){9-15} 
                          & AR@1   & AR@2   & AR@3   & AR@4   & AR@5   & AR@10  & AR@1\%  & AR@1   & AR@2   & AR@3   & AR@4   & AR@5   & AR@10  & AR@1\%  \\ \midrule
M2DP \cite{RN50}                      & 0.4684 & 0.5145 & 0.5563 & 0.5901 & 0.6345 & 0.8041 & 0.8517 & 0.7263 & 0.7536 & 0.776  & 0.7853 & 0.7931 & 0.8533 & 0.9019 \\
IRIS \cite{RN30}                     & 0.4718 & 0.5531 & 0.6091 & 0.6604 & 0.6979 & 0.8751 & 0.9205 & 0.7714 & 0.8081 & 0.837  & 0.8519 & 0.8641 & 0.9356 & 0.9632 \\
CSSC \cite{RN51}                     & 0.4743 & 0.5313 & 0.5864 & 0.6351 & 0.6644 & 0.8578 & 0.9034 & 0.7542 & 0.7851 & 0.8103 & 0.8311 & 0.8433 & 0.9103 & 0.9513 \\
SC \cite{RN11}                        & 0.4771 & 0.5534 & 0.6021 & 0.6634 & 0.6913 & 0.8801 & 0.9245 & 0.7451 & 0.7741 & 0.7954 & 0.8175 & 0.8217 & 0.9051 & 0.9410  \\
OT \cite{RN20}                        & 0.4852 & 0.5352 & 0.5914 & 0.6478 & 0.6832 & 0.8694 & 0.9155 & 0.7594 & 0.7905 & 0.8148 & 0.8385 & 0.8492 & 0.9143 & 0.9532 \\
CVTNet \cite{RN44}                   & 0.4901 & 0.5563 & 0.6104 & 0.6704 & 0.7088 & 0.8892 & 0.9352 & 0.7638 & 0.7982 & 0.8203 & 0.8397 & 0.8552 & 0.9281 & 0.9602 \\
ISC \cite{RN31}                       & 0.4926 & 0.5621 & 0.6235 & 0.6761 & 0.7114 & 0.8941 & 0.9391 & 0.7693 & 0.8012 & 0.8242 & 0.8401 & 0.8518 & 0.9259 & 0.9564 \\
Ours                      & \textbf{0.5460}  & \textbf{0.6349} & \textbf{0.6952} & \textbf{0.7302} & \textbf{0.7871} & \textbf{0.9333} & \textbf{0.9731} & \textbf{0.8244} & \textbf{0.8651} & \textbf{0.8974} & \textbf{0.9184} & \textbf{0.9291} & \textbf{0.9664} & \textbf{0.9814} \\ \bottomrule
\end{tabular}%
}
\end{table*}

\begin{table}[h]
\centering
\caption{Forward and Reverse Loop Closure Information in JLU Campus Dataset Sequences.}
\label{table3}
\scalebox{0.75}{
\begin{tabular}{cccccccc}
\hline
                                          &                                             & \multicolumn{3}{c}{Forward Loop}                                                                                                                   & \multicolumn{3}{c}{Reverse Loop}                                                                                              \\ \cline{3-8} 
\multirow{-2}{*}{Sequence}                & \multirow{-2}{*}{Scans}                     & No.                                    & Initial frames                                      & Loop frames                                 & No.                                    & Initial frames                                      & Loop frames            \\ \hline
\multicolumn{1}{c|}{}                     & \multicolumn{1}{c|}{}                       & \multicolumn{1}{c|}{}                    & \multicolumn{1}{c|}{0000 - 0427}                   & \multicolumn{1}{c|}{1680 - 1974}                   & \multicolumn{1}{c|}{}                    & \multicolumn{1}{c|}{0000 - 0017}                   & 0052 - 0069                   \\
\multicolumn{1}{c|}{\multirow{-2}{*}{01}} & \multicolumn{1}{c|}{\multirow{-2}{*}{3951}} & \multicolumn{1}{c|}{\multirow{-2}{*}{2}} & \multicolumn{1}{c|}{1855 - 1870}                   & \multicolumn{1}{c|}{3722 - 3737}                   & \multicolumn{1}{c|}{\multirow{-2}{*}{2}} & \multicolumn{1}{c|}{0314 - 0330}                   & 3726 - 3744                   \\ \hline
\multicolumn{1}{c|}{02}                   & \multicolumn{1}{c|}{3056}                   & \multicolumn{1}{c|}{1}                   & \multicolumn{1}{c|}{0000 - 0105}                   & \multicolumn{1}{c|}{2998 - 3056}                   & \multicolumn{1}{c|}{0}                   & \multicolumn{1}{c|}{\cellcolor[HTML]{EFEFEF}}      & \cellcolor[HTML]{EFEFEF}      \\ \hline
\multicolumn{1}{c|}{}                     & \multicolumn{1}{c|}{}                       & \multicolumn{1}{c|}{}                    & \multicolumn{1}{c|}{}                              & \multicolumn{1}{c|}{}                              & \multicolumn{1}{c|}{}                    & \multicolumn{1}{c|}{0095 - 0123}                   & 1513 - 1528                   \\
\multicolumn{1}{c|}{}                     & \multicolumn{1}{c|}{}                       & \multicolumn{1}{c|}{}                    & \multicolumn{1}{c|}{}                              & \multicolumn{1}{c|}{}                              & \multicolumn{1}{c|}{}                    & \multicolumn{1}{c|}{1771 - 1804}                   & 2885 - 2916                   \\
\multicolumn{1}{c|}{}                     & \multicolumn{1}{c|}{}                       & \multicolumn{1}{c|}{}                    & \multicolumn{1}{c|}{\multirow{-3}{*}{0000 - 0010}} & \multicolumn{1}{c|}{\multirow{-3}{*}{0052 - 0060}} & \multicolumn{1}{c|}{}                    & \multicolumn{1}{c|}{1497 - 1538}                   & 3129 - 3171                   \\
\multicolumn{1}{c|}{}                     & \multicolumn{1}{c|}{}                       & \multicolumn{1}{c|}{}                    & \multicolumn{1}{c|}{}                              & \multicolumn{1}{c|}{}                              & \multicolumn{1}{c|}{}                    & \multicolumn{1}{c|}{1413 - 1432}                   & 3247 - 3278                   \\
\multicolumn{1}{c|}{}                     & \multicolumn{1}{c|}{}                       & \multicolumn{1}{c|}{}                    & \multicolumn{1}{c|}{}                              & \multicolumn{1}{c|}{}                              & \multicolumn{1}{c|}{}                    & \multicolumn{1}{c|}{1385 - 1397}                   & 3318 - 3335                   \\
\multicolumn{1}{c|}{}                     & \multicolumn{1}{c|}{}                       & \multicolumn{1}{c|}{}                    & \multicolumn{1}{c|}{\multirow{-3}{*}{0066 - 0111}} & \multicolumn{1}{c|}{\multirow{-3}{*}{3124 - 3151}} & \multicolumn{1}{c|}{}                    & \multicolumn{1}{c|}{0732 - 0862}                   & 3873 - 3997                   \\
\multicolumn{1}{c|}{}                     & \multicolumn{1}{c|}{}                       & \multicolumn{1}{c|}{}                    & \multicolumn{1}{c|}{}                              & \multicolumn{1}{c|}{}                              & \multicolumn{1}{c|}{}                    & \multicolumn{1}{c|}{0671 - 0711}                   & 4021 - 4070                   \\
\multicolumn{1}{c|}{}                     & \multicolumn{1}{c|}{}                       & \multicolumn{1}{c|}{}                    & \multicolumn{1}{c|}{}                              & \multicolumn{1}{c|}{}                              & \multicolumn{1}{c|}{}                    & \multicolumn{1}{c|}{0441 - 0509}                   & 4242 - 4305                   \\
\multicolumn{1}{c|}{\multirow{-9}{*}{03}} & \multicolumn{1}{c|}{\multirow{-9}{*}{4633}} & \multicolumn{1}{c|}{\multirow{-9}{*}{3}} & \multicolumn{1}{c|}{\multirow{-3}{*}{1517 - 1595}} & \multicolumn{1}{c|}{\multirow{-3}{*}{4557 - 4633}} & \multicolumn{1}{c|}{\multirow{-9}{*}{9}} & \multicolumn{1}{c|}{0088 - 0161}                   & 4531 - 4573                   \\ \hline
\multicolumn{1}{c|}{}                     & \multicolumn{1}{c|}{}                       & \multicolumn{1}{c|}{}                    & \multicolumn{1}{c|}{0079 - 0154}                   & \multicolumn{1}{c|}{0149 - 0226}                   & \multicolumn{1}{c|}{}                    & \multicolumn{1}{c|}{}                              &                               \\
\multicolumn{1}{c|}{\multirow{-2}{*}{04}} & \multicolumn{1}{c|}{\multirow{-2}{*}{5831}} & \multicolumn{1}{c|}{\multirow{-2}{*}{2}} & \multicolumn{1}{c|}{0792 - 1507}                   & \multicolumn{1}{c|}{4036 - 4806}                   & \multicolumn{1}{c|}{\multirow{-2}{*}{1}} & \multicolumn{1}{c|}{\multirow{-2}{*}{0000 - 0017}} & \multirow{-2}{*}{0051 - 0069} \\ \hline
\end{tabular}%
}
\end{table}

\begin{figure}[ht]
\centering
\includegraphics[width=3.4in]{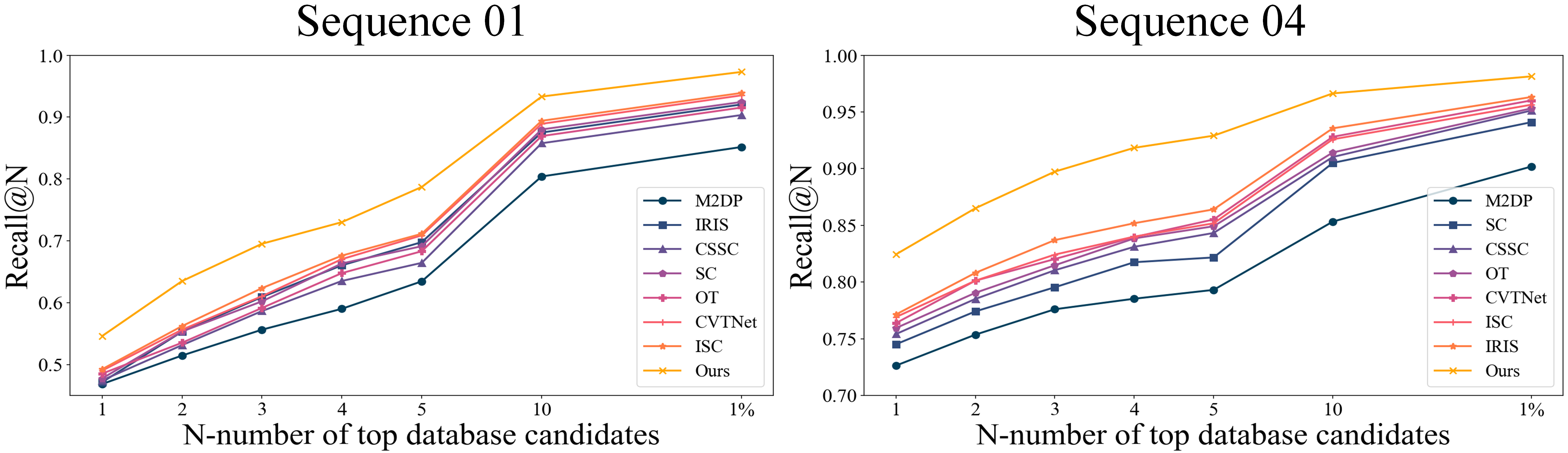}
\caption{Comparison of place recognition results on sequence 01 and 04 of JLU Campus dataset.}
\label{fig_8}
\end{figure}

Fig. \ref{fig_8} and Table \ref{Table4} showcase our method's performance on the JLU Campus dataset in both qualitative and quantitative aspects compared to other methods. Obviously, our method shows outstanding leading performance on both sequences. On sequence 01, the AR@1 for the OverlapTransformer is merely 0.4852, falling below the traditional method ISC’s 0.4926. Moreover, it performs worse than the latter on other metrics. This is mainly because, under the interference of movable objects, range image is more easily affected by occlusion. Range image-based approaches tend to lose more information compared to point cloud data. Our method consistently outperforms other algorithms in this complex scenario. Notably, we achieve an AR@1 score of 0.546, a significant 10.89\% improvement over the second-ranked ISC. Furthermore, with an AR@1\% score of 0.9731, it demonstrates the effectiveness of Range Transformer Module (RTM) in mitigating the impact of movable objects in challenging environments. The performance on sequence 04 further validates our method. The AR@1 of OverlapTranformer is 0.7594 on sequence 04, lower than ISC's 0.7693 and IRIS's 0.7714. Our method achieves 0.8244, showing an 8.56\% improvement over OverlapTransformer. It demonstrates the reliability of our method in environments with movable objects, showcasing its potential for practical autonomous driving applications.

\subsection{Ablation Study}

This ablation study validates the effectiveness of the Circular Convolution Module (CCM), the Range Transformer Module (RTM), and the Ovserlap-based loss function. All these experiments are conducted on sequence 02 of the KITTI dataset. In the experiments, we replaced CCM with a standard convolution module, removed RTM, and substituted the Overlap-based loss function with the original distance loss function to establish a baseline.

\begin{table}[h]
\centering
\caption{The ablation study of proposed module on sequence 02 of KITTI.}
\label{Table5}
\begin{tabular}{@{}cccccc@{}}
\toprule
\multirow{2}{*}{Approach} & \multicolumn{3}{c}{Module} & \multirow{2}{*}{\begin{tabular}[c]{@{}c@{}}Recall@1\end{tabular}} & \multirow{2}{*}{\begin{tabular}[c]{@{}c@{}}Recall@1\%\end{tabular}} \\ \cmidrule(lr){2-4}
                          & CCM     & RTM    & LOSS    &                                                                      &                                                                       \\ \midrule
\multirow{4}{*}{CCTNet}   & $\times$       & $\times$      & $\times$       & 0.738                                                                & 0.858                                                                 \\
                          & \checkmark       & $\times$      & $\times$       & 0.853                                                                & 0.947                                                                 \\
                          & \checkmark       & \checkmark      & $\times$        & 0.923                                                                & 0.971                                                                 \\
                          & \checkmark       & \checkmark      & \checkmark       & \textbf{0.968}                                                                & \textbf{0.994}                                                              \\ \bottomrule
\end{tabular}
\end{table}

\begin{figure}[h]
\centering
\includegraphics[width=3in]{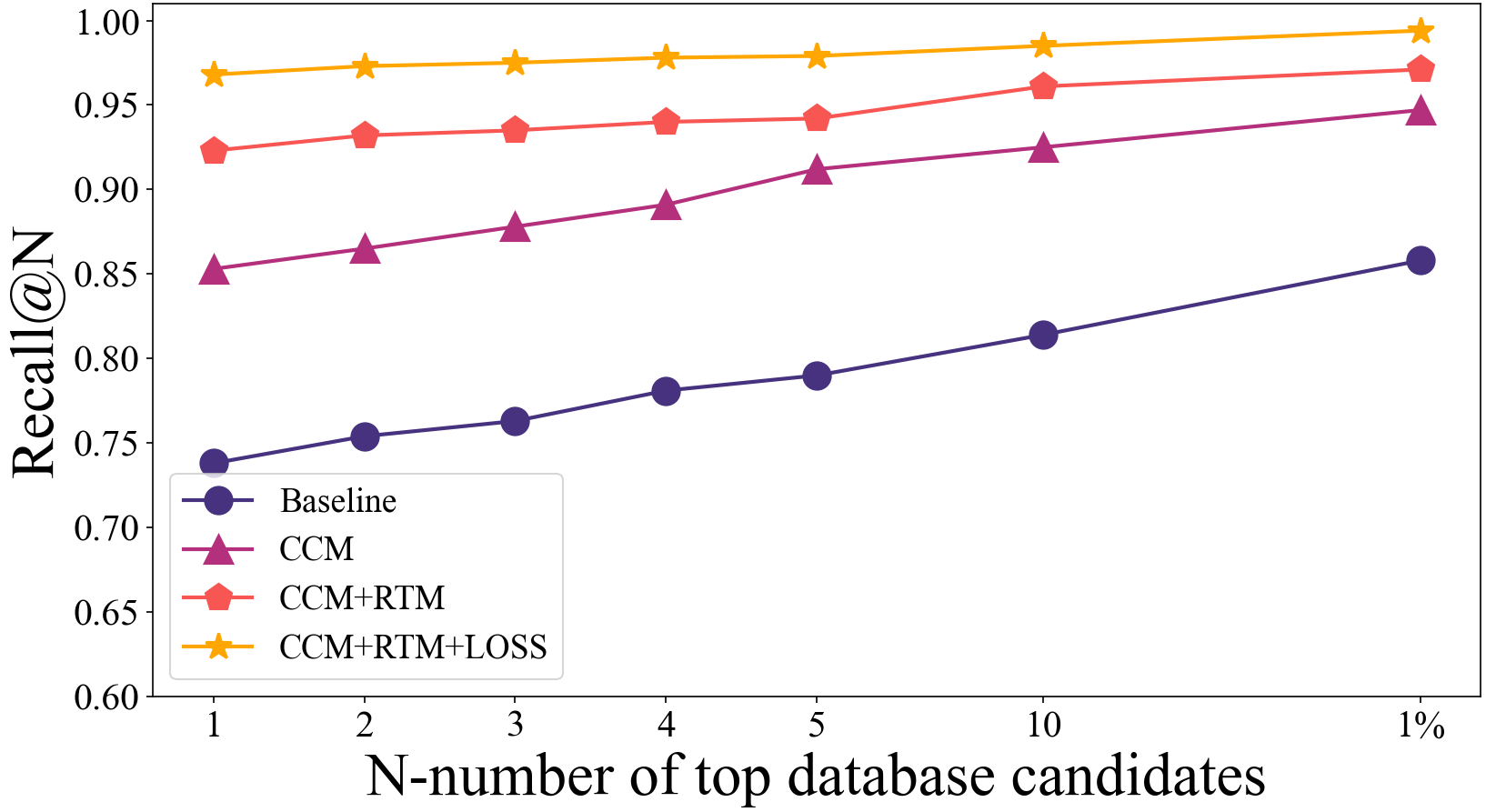}
 \caption{Ablation experiment of place recognition on sequence 02 of KITTI.}\vspace{-5mm}
\label{fig_9}
\end{figure}

As shown in Table \ref{Table5} and Fig. \ref{fig_9}, in the first experiment, introducing the CCM into baseline lead to an increase in Recall@1 from 0.738 to 0.853. It demonstrates that the CCM can enhances the network's performance by preserving point cloud structural features while extending the horizontal receptive field. In the second experiment, the addition of the RTM resulted in an 8.2\% improvement in Recall@1. It indicates that the combination of channel attention and spatial attention can effectively discerns movable objects from the background. In the final experiment, where both the CCM and the RTM are utilized along with the Overlap-based loss function. The Recall@1 score reaches 0.968, indicating that the Overlap-based loss function effectively guides the network in learning and enhances the descriptor's ability to represent complex scenarios.

\begin{figure*}[hb]
\centering
\includegraphics[width=6.8in]{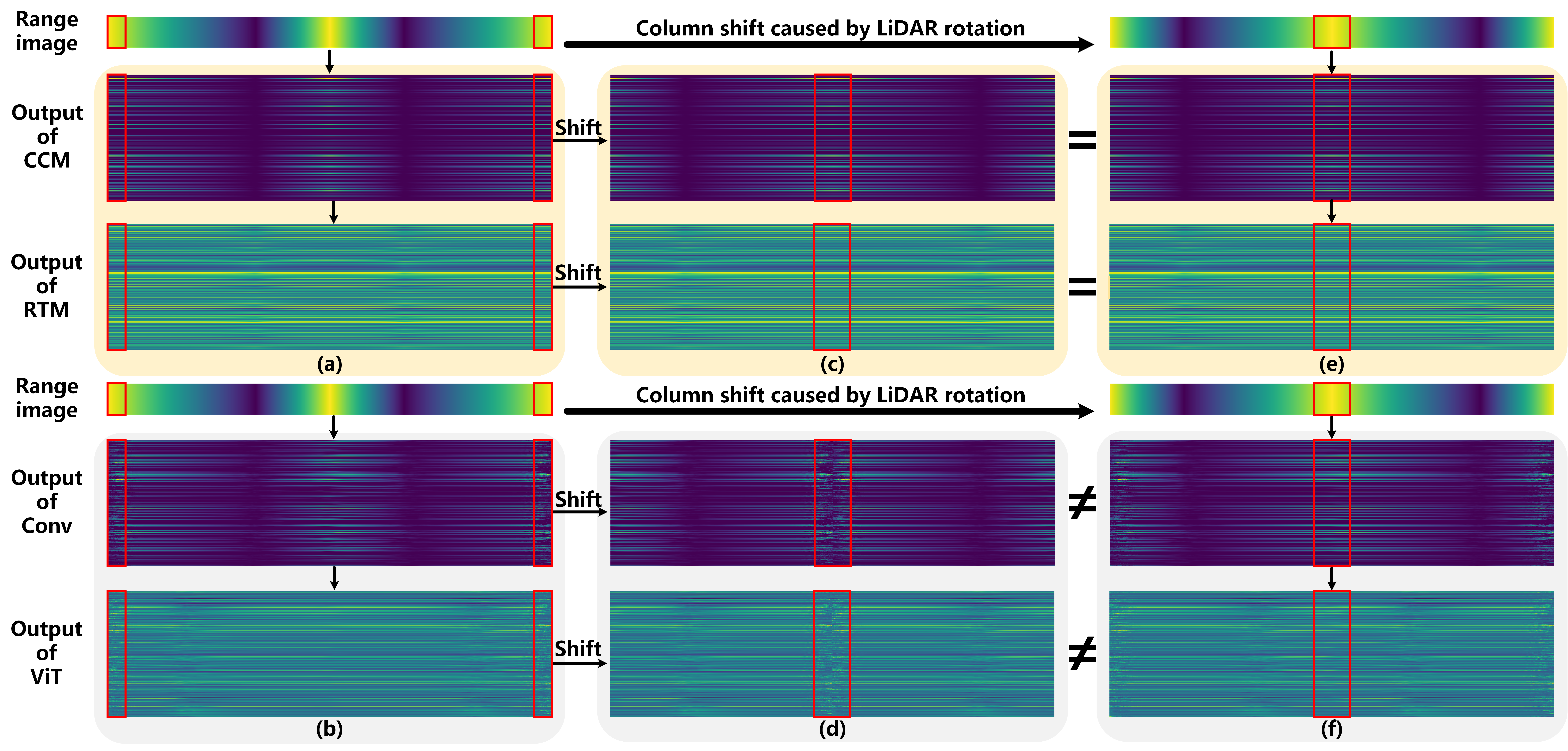}
\caption{An illustration of feature extraction stability under yaw angle changes exploiting a toy example. The feature map obtained through circular convolution and the transformer module, followed by a manual column shift, aligns with the results obtained from the image generated after LiDAR view-point change (as shown in (c) and (e)). The feature map generated by traditional convolution exhibits a crack, causing differences in the results (as depicted in (d) and (f)).}
\label{fig_10}
\end{figure*}

\subsection{Visualization Experiments of Circular Convolution}
In Sec. \ref{3b}, we have elaborated on CCM can effectively address the issue of inconsistent feature extraction caused by viewpoint changes. This section includes a comparative experiment to vividly demonstrate the described process. The feature maps generated by traditional convolution and circular convolution under viewpoint changes are shown in Fig. \ref{fig_10}. It also presents the impact of these results on subsequent Transformer modules. 

The main distinction between the results of circular convolution (see Fig. \ref{fig_10} (a)) and the traditional convolution (see Fig. \ref{fig_10} (b)) lies in the edge features at both ends of the image. Traditional convolution introduces blurry edge features due to zero-padding, while circular convolution preserves continuous and smooth edge features. Moreover, influenced by the convolutional results, the edge features from the traditional ViT module remain blurry, while the results from the RTM still smooth, continuous, and clear. We shifted the range image by half of its width to simulate a 180° rotation of the LiDAR viewpoint. The image is then input into the CCM and traditional convolution module yielding the results as shown in (e) and (f). Finally, we perform a half-width column-wise translation on the initial results (a) and (b), yielding results (c) and (d). By comparing (c) and (e), it is evident that the outputs of CCM and RTM are identical. However, the shift introduces gaps in the red box of (d) due to blurry features, distinct from the result in (f). It indicates that traditional convolution fails to recognize the panoramic nature of the range image, disrupting its spatial continuity. The experiment further verifies that CCM can maintain feature content when point cloud is rotated, which is crucial for generating rotation-invariant global descriptors and improving place recognition robustness.

\subsection{Runtime}
The runtime for each method is shown in Table \ref{Table6}. The running time is recorded for both descriptor extraction and searching steps. In the descriptor extraction step, Fast Histogram is the fastest method, while our method runs in 1.53ms, slightly inferior to Fast Histogram but much faster than the sampling speed of LiDAR (100 ms). In the descriptor search stage, our method only takes 0.39ms. It excels other methods, indicating that the descriptors generated by CCTNet are more discriminative. The above results show that our method can achieve real-time location recognition tasks, attributed to our lightweight network structure.

\begin{table}[h]
\centering
\caption{Comparison of runtime with state-of-the-art methods.}
\label{Table6}
\scalebox{0.90}{
\begin{tabular}{@{}cccc@{}}
\toprule
\multicolumn{2}{c}{\multirow{2}{*}{Approach}}               & Descriptor          & \multirow{2}{*}{Searching {[}ms{]}} \\ \cmidrule(lr){3-3}
\multicolumn{2}{c}{}                                        & Extraction {[}ms{]} &                                     \\ \midrule
\multirow{3}{*}{Hand crafted}   & Fast Histogram \cite{RN14}    & \textbf{1.07}                & 0.46                                \\
                                & Scan Context \cite{RN11}      & 57.95               & 492.63                              \\
                                & LiDAR Iris \cite{RN30}       & 7.13                & 9315.16                             \\ \midrule
\multirow{7}{*}{Learning based} & PointNetVLAD \cite{RN12}      & 13.87               & 1.43                                \\
                                & OverlapNet \cite{RN19}         & 4.85                & 3233.3                              \\
                                & NDT-Transformer-P \cite{RN18}  & 15.73               & 0.49                                \\
                                & MinkLoc3D \cite{RN16}          & 15.94               & 8.1                                 \\
                                & OT \cite{RN20} & 1.37                & 0.44                                \\
                                & CVTNet \cite{RN44}             & 15.03               & 0.44                                \\
                                & Ours                      & 1.53                & \textbf{0.39}                                \\ \bottomrule
\end{tabular}
}
\end{table}

\section{Conclusion}
In this paper, a LiDAR-based place recognition method was proposed which considers point cloud structural information and addresses movable object occlusion in the scenario. A circular convolution module was first introduced, which not only enlarges the network’s receptive field but also captures the ring-like structural features of the point cloud. In addition, a transformer module, integrating both channel and spatial attention, was employed to address the issue of excessive focus on a specific region by a single spatial attention in the presence of occlusions. Moreover, an overlap-based regression loss function was proposed to enhance the model's domain adaptation capability. The proposed method achieved Recall@1 scores of 0.924 and 0.965 on the KITTI and Ford Campus datasets, respectively, reaching the best level. To further evaluate the performance in scenarios with multiple movable objects, we conducted extensive comparison experiments on JLU Campus dataset. The experimental results suggest that our method outperforms the compared methods. Future work may aim to integrate temporal information into our current model.

\bibliographystyle{IEEEtran}
\bibliography{IEEEabrv,ref}

\begin{IEEEbiography}[{\includegraphics[width=1in,height=1.25in,clip,keepaspectratio]{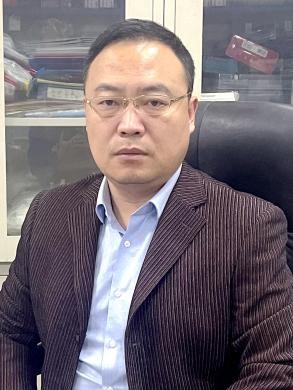}}]{Gang Wang}
received the Ph.D. degree in the oretical computer science from Jilin University (JLU). He is a professor and Ph.D. supervisor at the College of Computer Science and Technology, JLU. His research is closely centered on the industrial development trend. At the same time, he has been engaged in the research of intelligent software and equipment for a long time, including environmental perception, positioning, planning, and decision-making. In addition, based on the needs of industrialization, the theories of artificial intelligence are transformed and applied to robots, autonomous driving, unmanned vehicles, unmanned aerial vehicles, industrial control, security monitoring, sightseeing, patrol, reconnaissance, logistics transportation and other industries.
\end{IEEEbiography}

\begin{IEEEbiography}[{\includegraphics[width=1in,height=1.25in,clip,keepaspectratio]{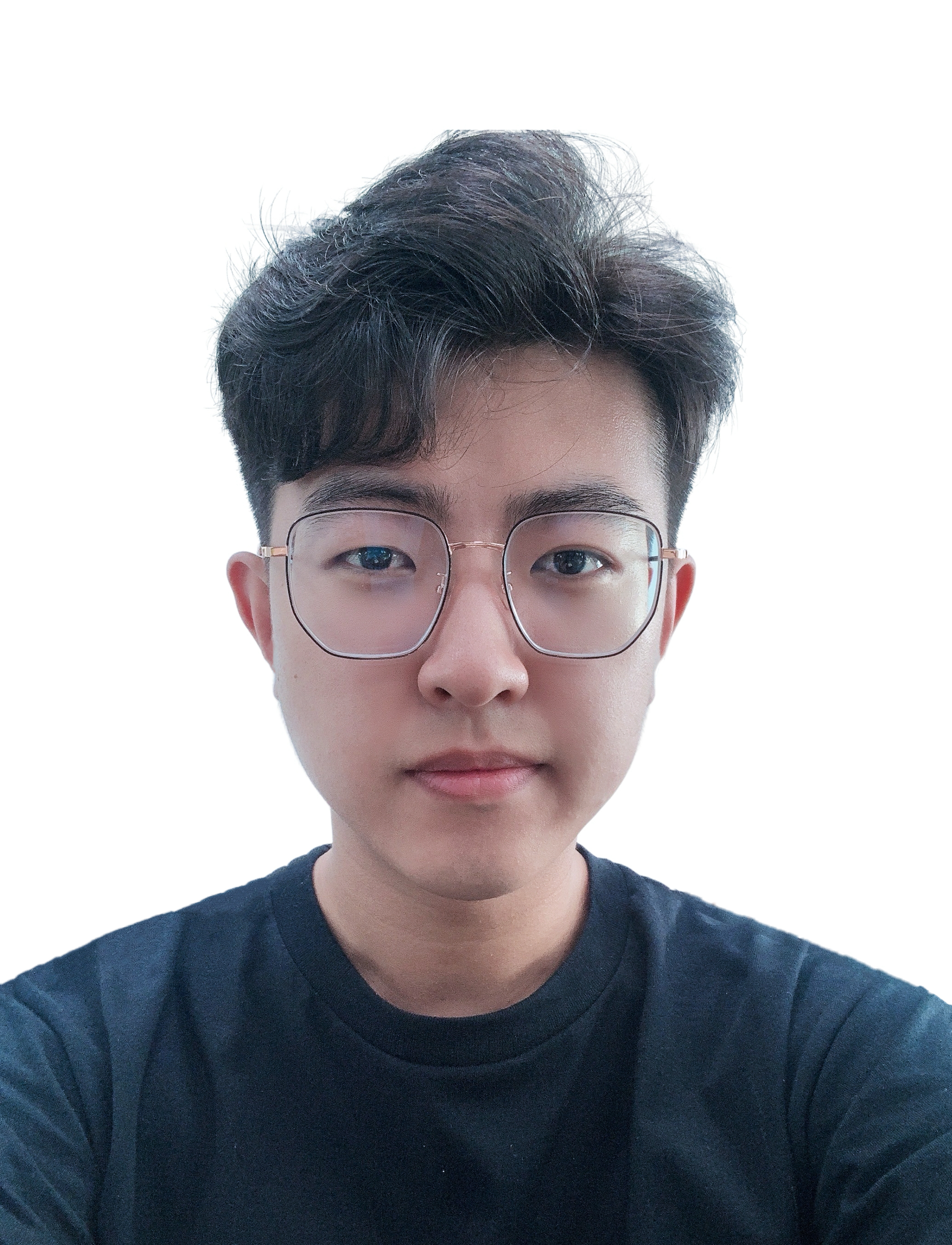}}]{Chaoran Zhu}
received the B.S. degree in Computer Science and Technology from Changchun University of Science and Technology, Changchun, China, in 2022. He is currently working toward a M.S. degree in Computer Science and Technology with Jilin University, Changchun, China. His current research interests include simultaneous localization and mapping and deep learning.\end{IEEEbiography}
\begin{IEEEbiography}[{\includegraphics[width=1in,height=1.25in,clip,keepaspectratio]{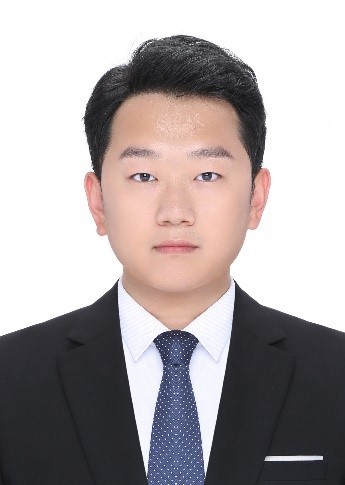}}]{Qian Xu}
received the B.S. degrees in software engineering and the Ph.D. degree in computer application technology from Jilin University. He is currently working at China North Vehicle Research Institute (NOVERI). His research interests include object detection, image segmentation, SLAM and vehicle drive control.\end{IEEEbiography}
\begin{IEEEbiography}[{\includegraphics[width=1in,height=1.25in,clip,keepaspectratio]{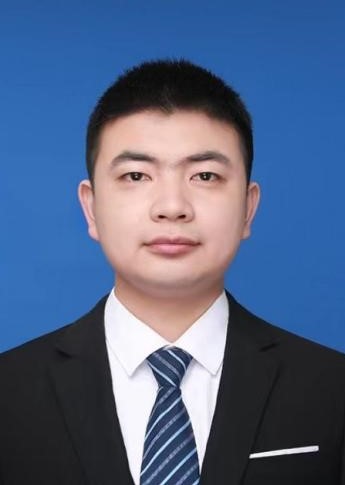}}]{Tongzhou Zhang}
received M.S. degrees from the College of Computer Science and Information Technology, Northeast Normal University, Changchun, China, in 2020. He is currently working toward a Ph.D. degree in software engineering with Jilin University, Changchun, China. His current research interests include simultaneous localization and mapping and mobile robotics.\end{IEEEbiography}
\begin{IEEEbiography}[{\includegraphics[width=1in,height=1.25in,clip,keepaspectratio]{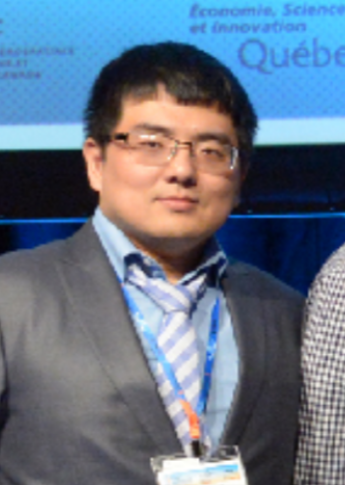}}]{Hai Zhang}
is a Full Professor at the Harbin Institute of Technology. He received his Ph.D. degree from Laval University, Canada in 2017 under the supervision of Prof. Xavier Maldague. Then, he was a postdoctoral research fellow at Laval University and University of Toronto with Prof. Andreas Mandelis and Prof. Shaker Meguid. He was also a visiting researcher in Fraunhofer EZRT, Fraunhofer IZFP and Technical University of Munich, Germany with Prof. Christian Grosse, etc.
His research interest includes infrared thermography, 3D photothermal coherent tomography, non-destructive evaluation of composite materials, non-invasive inspection of cultural heritage, industrial and medical imaging, etc. Based on his multi-field research background, he published over 100 papers in peer-review journals and international conferences, and won 3 Best Paper Awards. His H-index is 23 with more than 1300 citations and three ESI highly cited papers (from Scopus and Web of Science). His publications have been cited by many distinguished researchers in reputable journals including Nature Nanotechnology, etc.
He is also the Associate Editor of Measurement, Infrared Physics \& Technology, the editor of Quantitative InfraRed Thermography Journal, Russian Journal of Nondestructive Testing, Measurement Sensors, the guest editor of several scientific journals and the reviewer of more than 40 peer-review journals.
\end{IEEEbiography}
\begin{IEEEbiography}[{\includegraphics[width=1in,height=1.25in,clip,keepaspectratio]{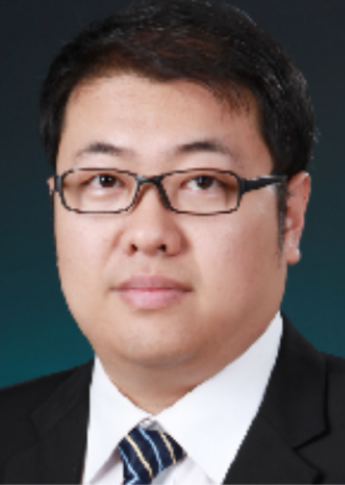}}]{Xiaopeng Fan}
(Senior Member, IEEE) received the B.S. and M.S. degrees from Harbin Institute of Technology (HIT), Harbin, China, in 2001 and 2003, respectively, and the Ph.D. degree from The Hong Kong University of Science and Technology (HKUST), Hong Kong, in 2009. From 2003 to 2005, he was a Software Engineer at Intel Corporation, China. In 2009, he joined HIT, where he is currently a Professor. From 2011 and 2012, he was a Visiting Researcher at Microsoft Re search Asia. From 2015 to 2016, he was a Research Assistant Professor at HKUST. Since 2018, he has been with Peng Cheng Laboratory. He has authored one book and over 180 articles in refereed journals and conference proceedings. His current research interests include video coding and transmission, image processing, and computer vision. He was a recipient of the Outstanding Contributions Award to the Development of IEEE Standard 1857 by IEEE in 2013. He served as the Program Chair for PCM2017, the Chair for IEEE SGC2015, and the Co-Chair for MCSN2015. He has been an Associate Editor of IEEE 1857 standard since 2012.\end{IEEEbiography}

\begin{IEEEbiography}[{\includegraphics[width=1in,height=1.25in,clip,keepaspectratio]{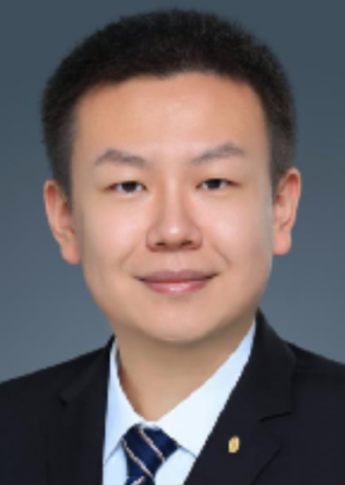}}]{Jue Hu}
is an Associate Professor at the Harbin Institute of Technology. He received his Ph.D. degree from University of Electronic Science and Technology of China in 2022 under the supervision of Prof. Guiyun Tian. He was also a visiting researcher in Laval University with Prof. Xavier Maldague. 
His current research interest includes infrared thermography, feature fusion framework and deep learning-based techniques for nondestructive imaging and evaluation. He was a recipient of the "Ermanno Grinzato – 5th Under 35 Years Best Paper Award" by AITA in 2021.
He is also the Guest Editor of Infrared Physics \& Technology, and serves as the reviewers of more than 10 peer-review journals.
\end{IEEEbiography}


\vfill

\end{document}